\documentclass[accepted]{article}

\usepackage{microtype}
\usepackage{graphicx}
\usepackage{subcaption}
\usepackage{booktabs} 
\usepackage{tcolorbox}
\usepackage{courier}
\usepackage{xcolor}
\usepackage{hyphenat}
\usepackage{framed}
\usepackage{courier} 
\usepackage{multirow}
\usepackage{multicol}
\usepackage{amsmath}
\usepackage{amssymb}
\usepackage{natbib}
\definecolor{promptbg}{gray}{0.90}
\colorlet{shadecolor}{promptbg}

\usepackage{hyperref}

\usepackage{media/icml2026}

\usepackage{mathtools}
\usepackage{amsthm}

\usepackage[capitalize,noabbrev]{cleveref}

\theoremstyle{plain}

\theoremstyle{definition}

\theoremstyle{remark}

\usepackage[textsize=tiny]{todonotes}

\icmltitlerunning{Recall as a Diagnostic for LLM Forecasting Errors}

\begin{document}

\twocolumn[
  \icmltitle{Approximate Recall, Approximate Forecasts: \\Recall as a Diagnostic for LLM Forecasting Errors}



    \icmlsetsymbol{equal}{*}

  \begin{icmlauthorlist}
    \icmlauthor{Prashanth Bhaskara}{equal,sch}
    \icmlauthor{Shubhaankar Gupta}{equal,sch}
    \icmlauthor{Seojoon Yeon}{equal,sch}
  \end{icmlauthorlist}

  \icmlaffiliation{sch}{University of Chicago}

  \icmlcorrespondingauthor{Prashanth Bhaskara}{pbhaskara@uchicago.edu}
  \icmlcorrespondingauthor{Shubhaankar Gupta}{shubhaankar@uchicago.edu}
  \icmlcorrespondingauthor{Seojoon Yoon}{seojoony@uchicago.edu}
  
  \icmlkeywords{Machine Learning, ICML}

  \vskip 0.3in
]
\printAffiliationsAndNotice{\icmlEqualContribution}  
\begin{abstract}
  Large Language Models (LLMs) exhibit approximate recall: they retain coarse associations about past events while losing fidelity on specific details. Past research on LLMs has quantified different models' forecasting ability in real-world events using prediction markets, but has not studied how losing said fidelity impacts their forecasting abilities. For example: while an LLM correctly recalls that the answer to: \textit{``Did Anna Sawai win the Best Drama Actress in the 76th Emmy Awards?''} is \textit{``Yes,"} it incorrectly states the event date as September 22 rather than September 15. \textbf{In this paper, we introduce a novel metric called the Recall Fidelity Score to study how imprecision in internalized knowledge propagates to forecasting capability on future events. On resolved historical markets, we find that higher local recall fidelity around semantically similar past events predicts lower forecasting error on resolved markets, and the same signal predicts smaller distance from open markets prices}.
  
\end{abstract}

\section{Introduction \& Related Work}
The emergence of reasoning capabilities in Large Language Models (LLMs) has made predictive reasoning a promising application domain for LLMs. Prophet Arena \cite{yangetal} introduced a Brier score-based benchmark for evaluating LLM forecasting accuracy on unresolved prediction market events. They discovered that frontier LLMs exhibit meaningful forecasting capability, but remain less calibrated than market-implied probabilities on Polymarket and Kalshi \cite{yangetal}.

Meanwhile, prior research such as \cite{machlab2024llm} and \cite{gemini-recall} illustrated that LLMs often exhibit ``approximate rather than precise recall'' of past events, particularly regarding exact dates and fine-grained details. The research indicated that relevant information is buried deep within context windows, leading to a ``needle in a haystack'' retrieval problem that limits reliable recall \cite{liu2023lost}. These findings raise the question of whether failures in recalling past events systematically degrade forecasting quality, and whether inaccurate recall leads models to produce overconfident but poorly calibrated predictions.

A key challenge is that forecasting metrics such as the Brier score conflate calibration errors with failures of underlying knowledge recall \cite{brier1950verification}. Two models may obtain similar Brier scores despite behaving very differently: one may remain cautious and weakly informative, while another may produce highly confident predictions grounded in inaccurate recollections of past events, analogous to the “imitative falsehoods” identified in prior work on LLMs \cite{lin2022truthfulqa}. In particular, confidently incorrect forecasts may suggest that faulty internalized knowledge distorts predictive reasoning. Motivated by these observations, we investigate how inaccuracies in LLM internalization and recall of past events affect forecasting performance on future events.

\section{Data and Methodology}

We study how a model's ability to recall past events within a domain $D \in$ \textit{\{Climate, Commodities, Companies, Economics, Elections, Entertainment, Financials, Politics, Science \& Technology\}} predicts its ability to forecast future events within and/or across domains. To construct the dataset, we collected binary prediction markets from Polymarket. Markets were passed through a multi-stage filtering pipeline described in \ref{app:dataset-construction}, then split along a knowledge-cutoff axis into two datasets: a recall set of resolved markets from \textit{2023-01-01} to \textit{2025-01-01}, and a forecast set of unresolved markets from \textit{2026-09-01} to \textit{2030-01-01}. The final corpus contains 1,300 recall markets and 1,350 forecast markets, for 2,650 unique markets.

We evaluate Claude Opus 4.7 \cite{anthropic2026claudeopus47}, Claude Sonnet 4.6 \cite{anthropic2026claudesonnet46}, GPT-5.5 \cite{openai2026gpt55}, and Gemini 3.1 Pro \cite{google2026gemini31pro}, representing frontier forecasting systems across distinct model families. This gives us a small but diverse set of high-performing models, allowing us to test whether recall–forecast relationships are consistent across architectures. We note that our design is observational -- recall and forecast quality may share common causes (e.g., training data coverage), and we cannot rule out this confound. We treat recall as a diagnostic signal: a model's pattern of recall errors on historically analogous events is a useful predictor of where its forecast errors will concentrate regardless of the underlying mechanism. Per-model coverage statistics and knowledge cutoff dates are reported in \ref{app:model-coverage}. Each $(\text{model}, \text{market})$ pair was queried exactly once with a shared system prompt and hyperparameters across all providers given in \ref{app:recall-prompt} and \ref{app:forecast-prompt}.

\subsection{Metrics}
We define metrics to assess model performance across different temporal contexts: recall before market resolution (pre-cutoff) and forecast accuracy after market resolution (post-cutoff).

\subsubsection{Recall Brier Score}
To evaluate accuracy in the pre-cutoff period, we compute the recall Brier Score

\begin{equation}
\label{eq:recall-brier}
\mathrm{Brier}_{\mathrm{recall}}
=
\overline{(p_t - y)^2},
\qquad
y \in \{0,1\}.
\end{equation}

where $p_t$ is the model's probability estimate and $y$ is the binary outcome of market $m$ on domain $D$. Lower values indicate better pre-cutoff forecasting accuracy.
 
\subsubsection{Recall Fidelity Score (RFS)}
We introduce the Recall Fidelity score, a new metric to study consistency in LLM recall accuracy by examining prediction fidelity during the pre-cutoff period:

\begin{equation}
\label{eq:recall-fidelity}
\begin{split}
\mathrm{RecallFidelity}_{m,d}(R_{m,d}) = \overline{s}_{m,d}, \\
s \in \{0, 0.5, 1\}
\end{split}
\end{equation}

We used a two-stage grading pipeline. First, each model's recalled outcome was compared against the true resolved market outcome using deterministic string matching. If the model could not recall, a value of \texttt{NULL} was assigned while incorrect recalls received $s=0.0$. If the recalled outcome was correct, the accompanying reasoning and evidence were fact-checked by with agentic web-search. The grader used a system prompt and parameters included in \ref{app:grader-prompt}.

\textbf{$s=0.0$}: The recalled outcome does not match the true resolved outcome.

\textbf{$s=0.5$}: The recalled outcome matches, but the reasoning or evidence contains clear factual errors.

\textbf{$s=1.0$}: The recalled outcome matches, and the reasoning and evidence are factually consistent.
 
\subsubsection{Forecast Squared Loss \& Log Loss (Brier Proxies)}
For unresolved forecast markets, true outcomes are unavailable, so we cannot compute post-resolution Brier score. As a proxy, we measure
disagreement between the model forecast and the contemporaneous market-implied probability:

\begin{equation}
\label{eq: fc-loss}
Loss_{fc}
=
\overline{(p_t - p_{m})^2}
\end{equation}
\begin{equation}
\label{eq: fc-log-loss}
LogLoss_{fc}
=
-\overline{
p_m \log(p_t)+(1-p_m)\log(1-p_t)
}
\end{equation}
where $p_t$ is the model's final probability estimate and $p_{m}$ is the consensus market probability.

\subsubsection{Local Recall Metric}
In order to measure recall more closely within specific domains, we construct a market-specific local recall score using Nearest Neighbors. Each market is represented as a TF--IDF \cite{salton1988term}, \cite{sparck1972statistical} vector over its question text and metadata (i.e category, domain, market slug, etc), and market similarity is computed through cosine similarity \cite{lloyd1982least}. For each forecast market $f$, we identify the $k$ nearest resolved recall set markets $\mathcal{N}_k(f)$, excluding any sharing $f$'s \texttt{event\_slug} to prevent leakage from sibling markets in distribution and perpetual market families, and compute the model's average recall score on those neighbors: \begin{equation}
R^{\text{local}}_{m,f,k}
=
\frac{1}{k}
\sum_{r \in \mathcal{N}_k(f)}
R_{m,r},
\end{equation}
where $R_{m,r} \in \{0, 0.5, 1\}$ is the graded recall score for model $m$ on resolved market $r$. The recall score measures how well a model's ability to recall nearby historical facts predicts its forecast quality on a new market, rather than relying only on coarse domain averages.

\section{Results}
\subsection{Recall Ability}
\begin{figure}[ht]
  \begin{center}
    \centerline{\includegraphics[width=\columnwidth]{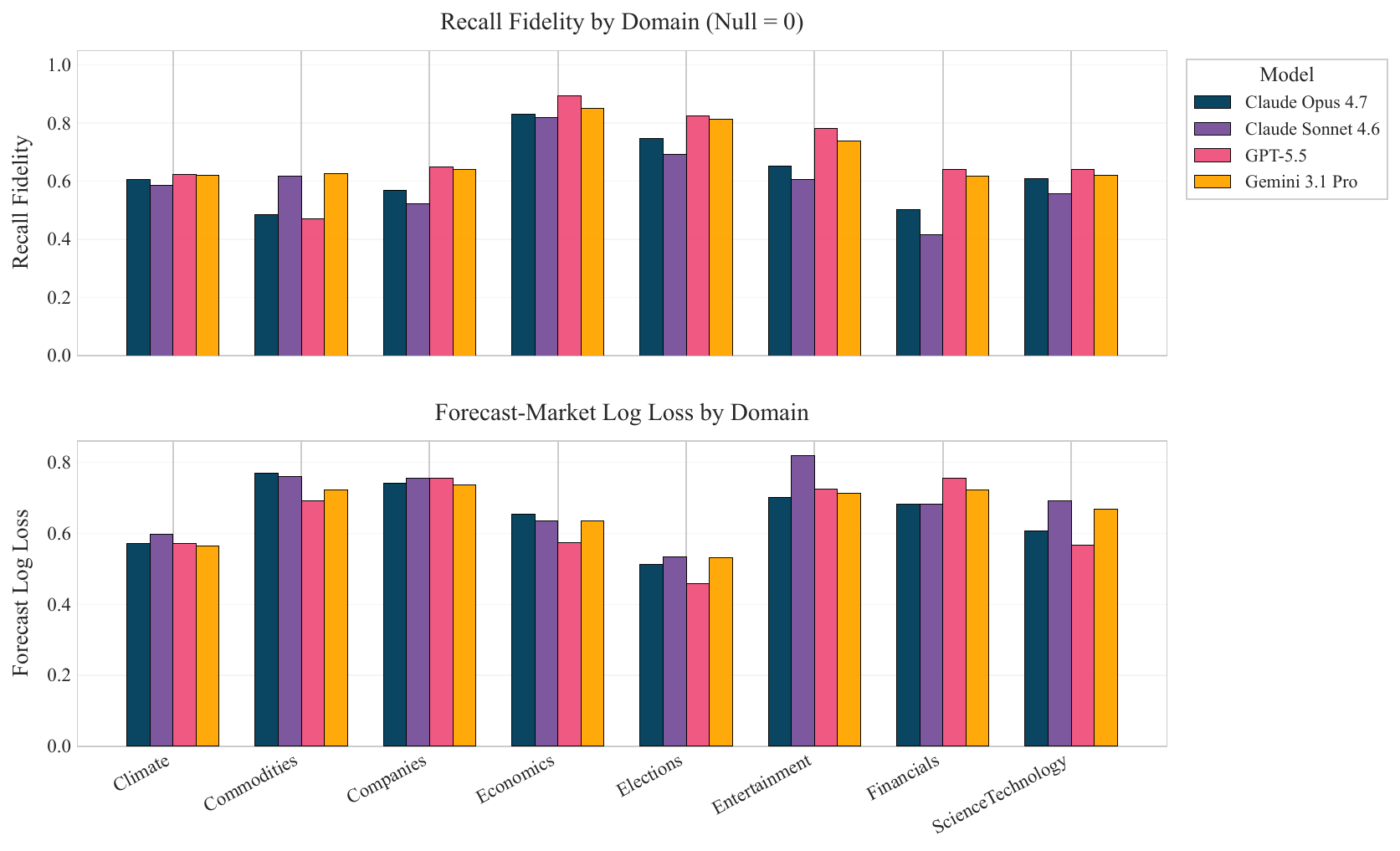}}
    \caption{Recall Fidelity Scores (null = 0) and Log Loss for all models grouped by domain. Elections and Economics domains seem to produce strongest recall
    across models. Exact figures reported in \ref{app:modelwise-domainwise}
      }
    \label{fig:domainwise-recall-forecast}
  \end{center}
\end{figure}

We compute domain-wise Recall Fidelity Score (Eq.\ref{eq:recall-fidelity}), $R_{m, d}$, and Brier score (Eq.\ref{eq:recall-brier}) for each model (\ref{app:modelwise-domainwise}). The RFS rewards clean reasoning regardless of how the model expressed its uncertainty in probability terms while the recall-task Brier rewards calibrated probabilistic judgement. Domain-wise RFS and recall Brier score for each model are shown in Fig. \ref{fig:domainwise-recall-forecast} and \ref{app:modelwise-domainwise}.

Table \ref{tab:rfs-defn} validates RFS as a proxy for recall Brier: $\beta_{R_{m,d}}=-0.22^{***}$ is stable across bivariate (M1), difficulty-controlled (M2), and fixed-effect (M3) specifications. A shift from the lowest- to highest-recall domain-model cell (0.40 to 0.87 RFS) predicts $\approx 0.10$ Brier reduction, the spread between best and worst model in our panel.

\begin{table}[t]

  \caption{Domain-level regression of recall Brier scores on Recall Fidelity Score $R_{m,d}$ across $N=36$ (model, domain) cells. Entropy defined in \ref{app:entropy_defn}}
  \label{tab:domain-regression}
  \begin{center}
  \begin{small}
  \setlength{\tabcolsep}{5pt}
  \begin{tabular}{lccr}
    \toprule
    Specification & Intercept & $\beta_{R_{m,d}}$ & $R^2$ \\
    \midrule
    M1: $R_{m,d}$ & $0.226^{***}$ & $-0.204^{***}$ & 0.398 \\
    M2: $+D_{\text{entropy}}$ & $0.286^{***}$ & $-0.222^{***}$ & 0.429 \\
    M3: $+\text{Model FE}$ & $0.285^{***}$ & $-0.223^{***}$ & 0.691 \\
    \bottomrule
  \end{tabular}
  \end{small}
  \end{center}
  \footnotesize{$^{*}p<0.05$, $^{**}p<0.01$, $^{***}p<0.001$; $\beta \approx -0.22^{***}$}
  \label{tab:rfs-defn}
\end{table}

\begin{figure}[ht]
  \begin{center}
    \centerline{\includegraphics[width=\columnwidth]{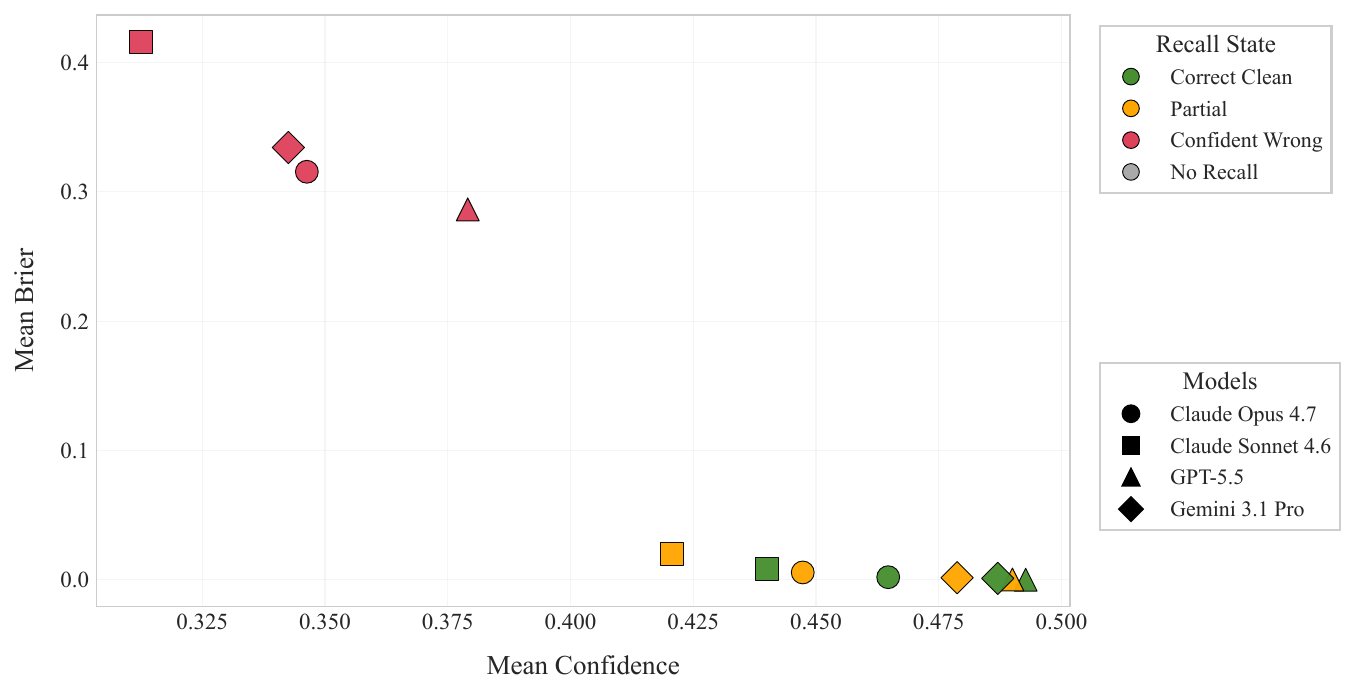}}
    \caption{Confident-wrong recall produces high Brier scores despite reduced confidence, while correct and partial recall yield near-zero Brier at high confidence. Confident-wrong cells are separated from all other states by an order of magnitude in Brier.
    }
    \label{fig: confidence-brier}
  \end{center}
\end{figure}

\subsection{Recall Decomposition \& Forecast Across Domains}
\subsubsection{Confidence Score}
Since the Recall Fidelity Score $R_{m,d} = \overline{s}, s \in \{0, 0.5, 1\}$ where $s$ is the grade assigned to an individual recall from the LLM, we can study the impact of a confidently wrong recall, where the model reports a remembered outcome that does not match the true resolution. Using the LLM generated probability $p$ for recall event, we define $confidence = |p-0.5|$ and plot it against the confidence levels for all models in Fig.~\ref{fig: confidence-brier}. We can see the visual asymmetry directly: confident-wrong cells (red) cluster at high Brier with intermediate confidence, while correct-clean and partial cells (green, yellow) cluster at near-zero Brier with high confidence.

\subsubsection{Local Recall}
We estimate the relationship between local recall and forecast performance using market level OLS regressions with model fixed effects and forecast market fixed effects, clustering standard errors by condition ID:

\begin{equation}
L_{m,f}
=
\beta R^{\text{local}}_{m,f,k}
+
\alpha_m
+
\gamma_f
+
\varepsilon_{m,f},
\label{eq:local_recall_reg}
\end{equation}

where $L_{m,f}$ is the model's forecast loss on the post-cutoff market. The main specification ($k=10$)  yields $\beta=-0.0652$ ($SE=0.0231$, $p=0.0048$, $N=4{,}897$, 1,264 clusters), indicating that models with stronger recall on nearby historical markets have lower forecast loss on similar forecast markets. These results remain robust across neighborhood sizes: the coefficient remains negative for $k \in \{3,5,10,20,50\}$. Negative control variants using shuffled, farthest neighbor, and cross-domain neighbors do not reproduce the same relationship, suggesting that the signal comes from local historical similarity.

We also report a domain-level robustness specification with model fixed effects and a domain-difficulty control, where difficulty is defined as the mean binary entropy of Polymarket prices within each domain; full specifications and estimates are provided in ~\ref{app:regression-specifications}.

\subsubsection{Recall Fidelity to Forecast}

\begin{figure}[ht]
  \begin{center}
    \centerline{\includegraphics[width=\columnwidth]{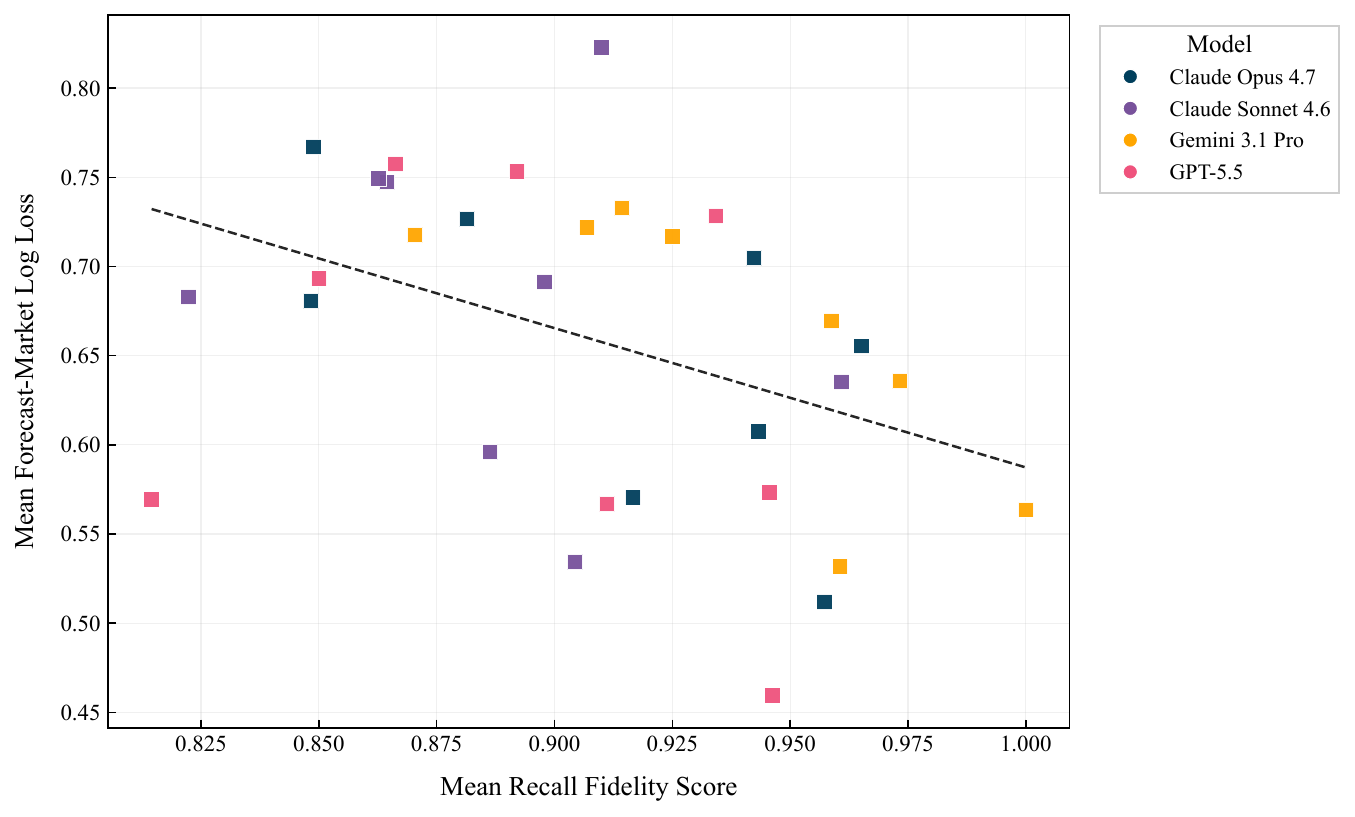}}
    \caption{Mean forecast log-loss (from unresolved markets) versus mean Recall Fidelity color coded by model across nine domains.}
    \label{fig: log-loss-vs-recall}
  \end{center}
\end{figure}

We provide model-wise, domain-wise Recall Fidelity Score, recall Brier, and Forecast losses in \ref{app:modelwise-domainwise}. Across the four models, we observe positive correlations between recall and forecast abilities, the magnitude of which varies by model and domain, as can be ascertained by Fig.\ref{fig:domainwise-recall-forecast} and Fig.\ref{fig: log-loss-vs-recall}. Upon directly regressing forecast log-loss on RFS, we find $\beta_{RFS} = -0.78$ with $p = 0.012^{*}$, pointing to an inverse relationship between the two.

\subsection{Local Recall Error-Type Decomposition.}
We further decompose local recall to distinguish whether forecast degradation is driven by missing knowledge, partially correct recall, or confidently incorrect recall. For each forecast market $f$, we compute the model's ability to recall historical facts of $k=10$ nearest resolved recall markets (Eq.\ref{eq:local_recall_reg}). We regress forecast loss on these local error-type shares with model and forecast-market fixed effects, using the partial-recall share as the omitted reference category.

\begin{figure}[ht]
  \begin{center}
    \centerline{\includegraphics[width=\columnwidth]{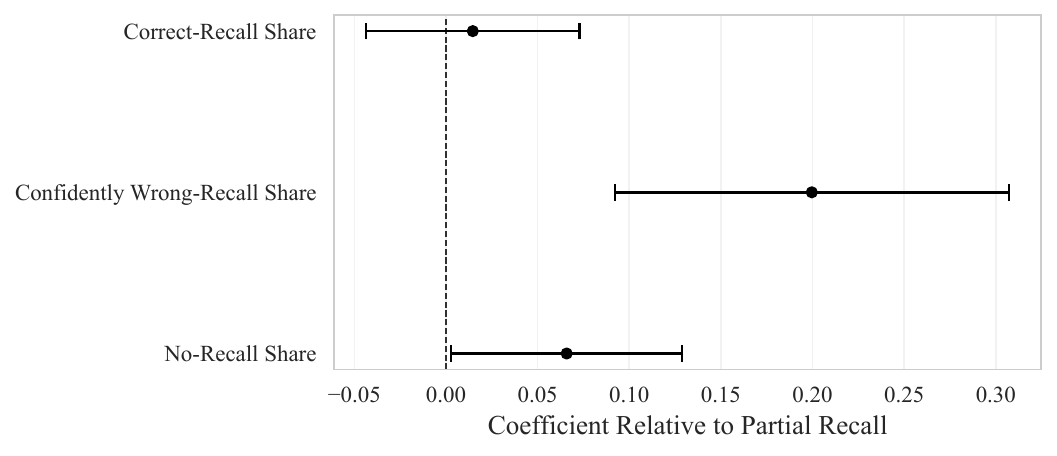}}
    \caption{Local recall error-type decomposition. Coefficients and 95\% confidence intervals come from a market-level OLS regression of forecast Brier loss on the shares of the ten nearest historical recall markets falling into each recall category, with model and forecast-market fixed effects and standard errors clustered by condition ID.}
    \label{fig: localrecalldecomp}
  \end{center}
\end{figure}

Fig.\ref{fig: localrecalldecomp} shows that the strongest predictor of worse forecasting is local confident-wrong recall. The coefficient on the local wrong-recall share $\beta=0.1996$ ($SE=0.0549$, $p=0.0003$), meaning that forecasts are substantially worse when a model's nearest historical analogues are cases where it recalled the wrong outcome. The local no-recall share is also associated with higher forecast loss, though more weakly ($\beta=0.0659$, $SE=0.0321$, $p=0.0405$). By contrast, the local correct-recall share is not statistically distinguishable from the partial-recall baseline ($\beta=0.0146$, $SE=0.0297$, $p=0.6223$). These results suggest an association between forecast degradation and locally relevant yet incorrectly remembered outcomes, rather than gaps in memory.

\subsection{Open-Market Forecast Proxy: Squared Distance and Soft-Target Log Loss}
\begin{figure}[ht]
  \begin{center}
    \centerline{\includegraphics[width=\columnwidth]{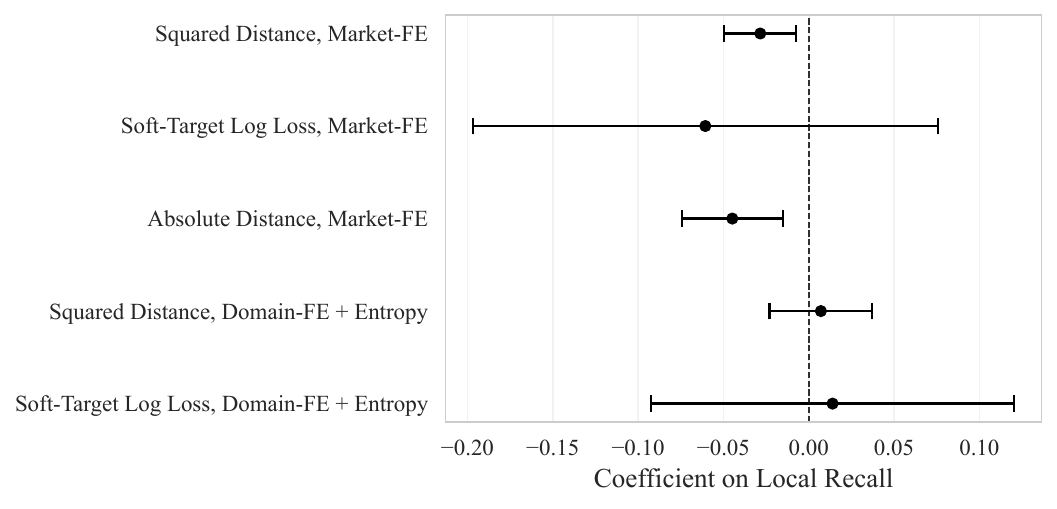}}
    \caption{Regression coefficients for local historical recall across open-market Brier-proxy specifications. Coefficient on $R^{\mathrm{local}}_{m,i,10}$ with 95\% confidence intervals using standard errors clustered by condition ID shown. Higher local recall significantly predicts lower squared forecast-market disagreement and lower absolute forecast-market distance.}
    \label{fig:open-proxy-plus-logloss-coefficients}
  \end{center}
\end{figure}

We estimate whether local historical recall predicts open-market forecast disagreement by regressing our aforementioned Brier proxies (Eq. \ref{eq: fc-loss}, \ref{eq: fc-log-loss}) on $R^{\mathrm{local}}_{m,i,10}$, the model's average recall score on the ten nearest historical resolved markets. For squared forecast-market disagreement, higher local recall is associated with lower disagreement: $\beta=-0.0284$ ($SE=0.0108$, $p=0.0084$, $N=5{,}724$, 1,897 clusters). The result is similar using absolute forecast-market distance ($\beta=-0.0448$, $SE=0.0152$, $p=0.0032$). Full historical and open-market regression specifications are provided in ~\ref{app:regression-specifications}. We also estimate domain-level specifications with model fixed effects, domain fixed effects, and a market-entropy control (\ref{app:entropy_defn}). Overall, the open-market proxy results suggest that local recall is most informative at the fine-grained market level, where historically analogous recall failures map onto specific forecast tasks. 

\section{Conclusion}

We find that LLM forecast quality is degraded by failures in knowledge internalization, with confidently wrong recall, not absent recall, emerging as the dominant driver of forecast-market disagreement across domains. Forecasting errors are systematically associated to the model's ability to faithfully recall nearby historical facts, and confident false recall remains the most damaging form of error. These results suggest that efforts to improve LLM forecasting should prioritize reducing false memory over expanding knowledge coverage, as a model that knows it doesn't know outperforms one that confidently misremembers.

\nocite{langley00}

\bibliography{citations}
\bibliographystyle{media/icml2026}

\newpage
\appendix
\onecolumn

\section{Appendix}
\label{app:prompts}

\subsection{Recall Prompt}
\label{app:recall-prompt}
We describe the exact prompts used to elicit recall judgments and 
probabilistic forecasts from each model. 

The recall prompt is presented to each model for every pre-cutoff resolved market. 

\begin{shaded}
\begin{ttfamily}
\small
\noindent\textbf{SYSTEM PROMPT:}\\
\noindent You are an AI assistant specialized in forecasting AND in recalling 
previously seen real-world events. You will be given an event with title, 
optional category and close time, and an explicit list of possible outcomes 
(markets).

\medskip
\noindent\textbf{Goal:} If you genuinely REMEMBER this exact event from your 
training data, USE that memory.

\medskip
\noindent\textbf{Instructions:}\\
1) Output strictly valid JSON (no extra text) with the exact schema below.\\
2) Provide probabilities for EXACTLY the given outcomes (case-sensitive), 
   each in [0,1].\\
3) If you remember the event, include concrete verifying details (e.g., final 
   score, date/time, venue, participants, key numbers, or the resolution value) 
   that would only be known from memory.\\
4) If you do NOT truly remember the event, set \texttt{recognized\_event=false} 
   and do NOT fabricate specifics. Give best-effort probabilities from 
   reasoning only.

\medskip
\noindent\textbf{Output JSON shape (exact keys):}

\medskip
\noindent\{\\
\hspace*{1em}``rationale'': ``One or two short sentences (max 50 words).'',\\
\hspace*{1em}``probabilities'': \{ ``\textless{}outcome1\textgreater{}'': 
  \textless{}float\textgreater{}, ``\textless{}outcome2\textgreater{}'': 
  \textless{}float\textgreater{}, \ldots \},\\
\hspace*{1em}``recall\_assessment'': \{\\
\hspace*{2em}``recognized\_event'': \textless{}true$|$false\textgreater{},\\
\hspace*{2em}``evidence\_facts'': [\\
\hspace*{3em}``Concrete verifying details you recall 
  (dates/scores/metrics/participants/\\\hspace*{4em}venue/etc.)'',\\
\hspace*{3em}``List at least 2 if recognized\_event=true; 
  otherwise leave empty''\\
\hspace*{2em}],\\
\hspace*{2em}``recalled\_outcome\_if\_known'': ``\textless{}verbatim outcome 
  name if you remember the resolution, else null\textgreater{}''\\
\hspace*{1em}\}\\
\}

\medskip
\noindent\textbf{Hard constraints:}\\
-- JSON only. No text before/after.\\
-- Use only the provided outcome names.\\
-- Do not invent specifics unless you genuinely remember them.

\medskip
\noindent\textbf{USER PROMPT:}\\
\noindent This is the event: \{title\}\\
Category: \{category\}\\
Close Time (UTC): \{close\_time\}

\medskip
\noindent Example market meaning (rules):\\
-- \{title\}: \{rule\_text\}

\medskip
\noindent Possible outcomes (provide probabilities for exactly these):\\
\{outcomes\_block\}

\medskip
\noindent Your JSON must look like:\\
\{\\
\hspace*{1em}``rationale'': ``\textless{}short 2--3 sentence rationale\textgreater{}'',\\
\hspace*{1em}``probabilities'': \{\\
\hspace*{2em}\{probs\_template\}\\
\hspace*{1em}\},\\
\hspace*{1em}``recall\_assessment'': \{\\
\hspace*{2em}``recognized\_event'': \textless{}true$|$false\textgreater{},\\
\hspace*{2em}``evidence\_facts'': [\\
\hspace*{3em}``\textless{}verifying detail 1\textgreater{}'',\\
\hspace*{3em}``\textless{}verifying detail 2\textgreater{}''\\
\hspace*{2em}],\\
\hspace*{2em}``recalled\_outcome\_if\_known'': ``\textless{}outcome name if you 
  remember the resolution, else null\textgreater{}''\\
\hspace*{1em}\}\\
\}

\end{ttfamily}
\end{shaded}

\subsection{Forecast Prompt}
\label{app:forecast-prompt}

The forecast prompt is presented to each model for every post-cutoff 
market. The model has no access to the true outcome and must reason 
from its internalized prior knowledge alone. We elicit a calibrated 
probability rather than a binary prediction.

\begin{shaded}
\begin{ttfamily}
\small
\noindent\textbf{SYSTEM PROMPT:}\\
\noindent You are an AI assistant specialized in analyzing and predicting 
real-world events.

\medskip
\noindent\textbf{Event:} \textless{}event title\textgreater{}\\
\textbf{Close Time (UTC):} \textless{}close\_time\textgreater{}

\medskip
\noindent\textbf{Example market rule:}\\
-- \textless{}market\_name\textgreater{}: \textless{}rule text\textgreater{}

\medskip
\noindent\textbf{Possible outcomes} (provide probabilities for exactly these):\\
-- \textless{}outcome\_1\textgreater{}\\
-- \textless{}outcome\_2\textgreater{}\\
-- \textless{}outcome\_3\textgreater{}\\
\hspace*{1em}\ldots

\medskip
\noindent\textbf{Constraints:}\\
1) Provide probabilities for exactly the listed outcomes (case-sensitive).\\
2) Do not invent additional outcomes.\\
3) Each probability must be a float in [0, 1].\\
4) Return JSON only; no extra text.

\medskip
\noindent\textbf{Output JSON:}

\medskip
\noindent\{\\
\hspace*{1em}``rationale'': ``\textless{}concise 2--3 sentence 
  rationale\textgreater{}'',\\
\hspace*{1em}``probabilities'': \{\\
\hspace*{2em}``\textless{}outcome\_1\textgreater{}'': 
  \textless{}float\textgreater{},\\
\hspace*{2em}``\textless{}outcome\_2\textgreater{}'': 
  \textless{}float\textgreater{},\\
\hspace*{2em}\ldots\\
\hspace*{1em}\}\\
\}

\medskip
\noindent\textbf{USER PROMPT:}\\
\noindent Here is the given event:\\
Event title: \{title\}\\
Category: \{category\}\\
Close time (UTC): \{close\_time\}

\medskip
\noindent Possible outcomes:\\
\{outcomes\_block\}

\medskip
\noindent Example rule excerpt: \{rule\_text\}

\end{ttfamily}
\end{shaded}

\subsection{Grader Prompt}
\label{app:grader-prompt}

The grader prompt is presented to Claude Haiku 4.5 (with web search, max
tokens 2000) for every recall item whose recalled outcome matches the actual
resolution under deterministic string matching. Items with mismatched
recalled outcomes are scored $0.0$ without invoking the grader.

\begin{shaded}
\begin{ttfamily}
\small
\noindent\textbf{SYSTEM PROMPT:}\\
\noindent You are a meticulous fact-checker. You will be given:\\
1. A prediction market question and its actual resolution\\
2. An LLM's reasoning explaining why it predicted the (correct) outcome

\medskip
\noindent\textbf{Your job:} identify factual errors in the reasoning. Check
concrete claims --- dates, names, years, ceremony numbers, chart positions,
box-office figures, who beat whom, song/album/film titles.

\medskip
\noindent\textbf{CRITICAL RULE ON FLAGGING:} Only flag a claim as wrong if
you are highly confident it is wrong. A confidently asserted ``correction''
that turns out to be wrong is worse than missing an error. If you are
pattern-matching from memory and not certain, do NOT flag it. When in
doubt, default to CORRECT. The \texttt{actual\_resolution} is your
strongest anchor --- claims consistent with it should not be flagged unless
clearly contradicted by other widely known facts.

\medskip
\noindent\textbf{Do NOT penalize:}\\
-- Hedging language (``I recall'', ``I think'', ``uncertain'')\\
-- Probabilistic estimates\\
-- Missing information (only flag stated facts that are wrong)\\
-- Minor rounding or approximations within $\sim$10\% (e.g. ``\$80M'' vs
   ``\$80.5M'', ``\$35M'' vs ``\$35.4M'')\\
-- Slight phrasing differences in titles, names, or roles

\medskip
\noindent\textbf{Flag as errors only when clearly wrong:}\\
-- Wrong dates or years (off by months/years, not days)\\
-- Wrong winners or who-beat-whom claims\\
-- Box-office or chart figures off by more than $\sim$10\%\\
-- Misattributed songs, albums, or films to the wrong artist\\
-- Internally inconsistent claims within the reasoning itself

\medskip
\noindent\textbf{Respond in this exact format:}

\medskip
\noindent VERDICT: CORRECT\\
or\\
VERDICT: ERRORS\_FOUND

\medskip
\noindent ERRORS:\\
-- \textless{}error 1, one line, with the correct fact\textgreater{}\\
-- \textless{}error 2, one line, with the correct fact\textgreater{}\\
(omit ERRORS section if VERDICT is CORRECT)

\medskip
\noindent\textbf{USER PROMPT:}\\
\noindent Question: \{title\}\\
Actual resolution: \{actual\_resolution\}\\
Close time: \{close\_time\}

\medskip
\noindent LLM's reasoning:\\
\texttt{"""}\\
\{rationale\}\\
\texttt{"""}

\medskip
\noindent Evidence facts the LLM cited:\\
\{evidence\_facts\}

\medskip
\noindent Fact-check the reasoning and evidence. Are there any factual
errors?

\end{ttfamily}
\end{shaded}

\subsection{Dataset Construction and Filtering Pipeline}
\label{app:dataset-construction}

All markets were collected from Polymarket's public Gamma API
(\url{https://gamma-api.polymarket.com}) using per-domain wrapper scripts
that delegate to a shared downloader. Two collection passes produce two
disjoint datasets along a knowledge-cutoff axis: the recall set
(\texttt{markets.csv}, \(\texttt{status=closed}\), close window
2023-01-01--2025-01-01) and the forecast set (\texttt{newmarkets.csv},
active markets, close window 2026-09-01--2030-01-01).

\medskip
\noindent\textbf{Eight-step filter chain.} Each candidate market returned
by the API must pass, in order, every check below; the first failure
discards the market. Counts in this section refer to the recall pull
unless noted.

\begin{shaded}
\begin{ttfamily}
\small
\noindent\textbf{Filter 1 --- Status.}\\
Recall pass: \texttt{status = "closed"}. Forecast pass: active markets
(\texttt{closed = false}, accepting outcome\hyp{}pending markets).

\medskip
\noindent\textbf{Filter 2 --- Resolved (recall only).}\\
\texttt{actual\_resolution} \(\in\) \{Yes, No\}. Markets with null,
ambiguous, or non-binary resolutions are dropped.

\medskip
\noindent\textbf{Filter 3 --- Close-time window.}\\
\texttt{close\_time} must lie strictly inside the requested
[\texttt{start\_date}, \texttt{end\_date}] interval (Unix-second
comparison after parsing the ISO-8601 timestamp).

\medskip
\noindent\textbf{Filter 4 --- Volume.}\\
If \texttt{--min-volume} is set, retain markets whose USD trade volume is
$\ge$ the threshold. The headline pull used the API default
(\texttt{min\_volume = None}); a published threshold will be added in v2.

\medskip
\noindent\textbf{Filter 5 --- Title-keyword exclusion.}\\
Drop any market whose title or slug matches any term in
\texttt{exclude\_keywords}. The forecast pass uses a fixed global list:\\
-- \texttt{"luigi mangione"}\\
-- \texttt{"manifesto"}\\
-- \texttt{"brian thompson"}\\
-- \texttt{"brian-thompson"}

\medskip
\noindent\textbf{Filter 6 --- Within-domain dedup.}\\
\texttt{condition\_id} (or \texttt{market\_slug} fallback) must not
already be in \texttt{seen\_ids} for this domain.

\medskip
\noindent\textbf{Filter 7 --- Cross-domain dedup.}\\
\texttt{condition\_id} must not appear in any previously written
\texttt{markets.csv} of another domain. This guarantees that no market
is counted in more than one domain.

\medskip
\noindent\textbf{Filter 8 --- First-N cap.}\\
Iterate the filtered stream until \texttt{len(markets) $\ge$ N}, where
$N=150$ for both the recall pass and the forecast pass (the forecast
script uses an upstream cap of 300 and is then truncated to 150).
Markets are accepted in API-default pagination order, not randomly
sampled.
\end{ttfamily}
\end{shaded}

\medskip
\noindent\textbf{Sampling procedure.} Sampling is non-random: the API
returns paginated results (\texttt{page\_limit = 100} per request,
\texttt{max\_pages = 200}) in its own internal ordering, and the
downloader accepts the first $N$ markets that pass Filters 1--7. There
is no stratification within a domain, no balancing across resolution
outcome, and no weighting by volume. The Climate domain produces only
100 resolved markets inside the 2023--2025 window that survive the
filter chain, so Climate's recall slice undershoots the $N=150$ target.

\medskip
\noindent\textbf{Deduplication rules.}
\begin{itemize}
\item \emph{Within-domain:} a market is identified by its
\texttt{condition\_id} (a 32-byte hex string assigned by Polymarket);
the \texttt{market\_slug} is used as a fallback only when
\texttt{condition\_id} is missing. Identical IDs from a later page are
ignored.
\item \emph{Cross-domain:} before each new domain pull, the downloader
loads every \texttt{markets.csv} already on disk and unions their
\texttt{condition\_id} columns into an exclusion set. Any candidate
matching this set is skipped, regardless of which tag returned it.
\item \emph{Same-event awareness for the local-recall analysis (downstream):}
markets sharing an \texttt{event\_slug} are not deduplicated at the
collection step but are excluded from each other's nearest-neighbor sets
in the market-level regression to prevent leakage.
\end{itemize}

\medskip
\noindent\textbf{API collection details.}
\begin{itemize}
\item Endpoint: \texttt{GET /markets} on
\texttt{https://gamma-api.polymarket.com}.
\item Query parameters: \texttt{limit=100}, \texttt{status},
\texttt{start\_time} / \texttt{end\_time} (Unix seconds), one
\texttt{tags=<slug>} parameter per domain tag, optional
\texttt{min\_volume}, optional free-text \texttt{search}.
\item Pagination: cursor-based, capped at 200 pages per (domain, tag)
pair.
\item Rate-limit: client-side throttle of 0.1\,s minimum between
requests; HTTP retries on 429/5xx with exponential backoff.
\item Recorded fields per market: \texttt{condition\_id},
\texttt{market\_slug}, \texttt{event\_slug}, \texttt{title},
\texttt{category}, \texttt{open\_time}, \texttt{close\_time},
\texttt{settlement\_ts}, \texttt{volume}, \texttt{status},
\texttt{actual\_resolution} (recall) or \texttt{market\_price}
(forecast).
\item Code: \texttt{scripts/polymarket\_market\_downloader.py} (recall
pass) and \texttt{scripts/download\_polymarket\_future\_markets.py}
(forecast pass).
\end{itemize}

\subsection{Domain Definitions and Market Taxonomy}
\label{app:domain-taxonomy}

The corpus is partitioned into nine disjoint domains. Domain membership
is determined entirely by Polymarket's own tag system; we do not
re-label markets manually. Each domain script supplies a fixed
\texttt{category} display name and a tuple of one or more Polymarket
tag slugs. A market is admitted to a domain if and only if the API
returns it for at least one of that domain's tag-keyed queries
\emph{and} it is not already in another domain's set (cross-domain
dedup, Filter 7 above).

\medskip
\noindent\textbf{Tag slugs per domain.} The tag tuple sent verbatim to
the Gamma API for each domain is:

\begin{shaded}
\begin{ttfamily}
\small
\noindent\textbf{Climate.}\\
\texttt{Weather, Climate, Global Temp, global warming, Hurricane,
Hurricanes}\\
(supplemental tag for under-fill: \texttt{science}, with keyword filter)

\medskip
\noindent\textbf{Commodities.}\\
\texttt{crypto, commodities}

\medskip
\noindent\textbf{Companies.}\\
\texttt{business, companies}

\medskip
\noindent\textbf{Economics.}\\
\texttt{economics}\\
(supplemental tags for under-fill: \texttt{business, finance, news,
current-events}, with keyword filter)

\medskip
\noindent\textbf{Elections.}\\
\texttt{elections}

\medskip
\noindent\textbf{Entertainment.}\\
\texttt{pop-culture, entertainment}

\medskip
\noindent\textbf{Financials.}\\
\texttt{finance, financials}

\medskip
\noindent\textbf{Politics.}\\
\texttt{politics, us-politics}

\medskip
\noindent\textbf{Science/Technology.}\\
\texttt{science, technology}
\end{ttfamily}
\end{shaded}

\medskip
\noindent\textbf{Domain mapping methodology.}

\begin{enumerate}
\item For each domain $d$, the API is queried with one
\texttt{tags=<slug>} parameter per primary tag in the tuple, paginated
as described in ~\ref{app:dataset-construction}.
\item Returned markets are passed through Filters 1--7. Cross-domain
dedup (Filter 7) operates in the order in which the per-domain scripts
are run; in practice this is alphabetical by output directory, so a
market that satisfies multiple tags is assigned to the first domain
that claims it.
\item If a domain's resolved-market pool falls short of $N=150$ after
the primary tags, a small \emph{supplemental} pull broadens the tag set
and applies a domain-specific keyword filter (\texttt{supplement\_kws}
in \texttt{fetch\_markets.py}) to keep the supplement topically aligned.
This affected only Climate (kept at 100, no supplement applied) and
Economics in the recall pass.
\item No manual relabeling is performed. The
\texttt{category} field stored on each market is purely a display
label; the operative grouping is the directory the market is written
to.
\end{enumerate}

\medskip
\noindent\textbf{Example markets.} Two illustrative recall-pass titles
per domain (drawn from \texttt{Research/<Domain>Markets/recall/}
\texttt{claude-opus-4-7\_1000\_recall.json}):

\begin{shaded}
\begin{ttfamily}
\small
\noindent\textbf{Climate.}\\
-- Iceland volcanic eruption by Nov 15?\\
-- Will the chopsticks catch SpaceX's Super Heavy?

\medskip
\noindent\textbf{Commodities.}\\
-- Will BTC or ETH reach all-time high first?\\
-- Will crude oil hit \$100 in 2023?

\medskip
\noindent\textbf{Companies.}\\
-- OpenSea acquired before March?\\
-- Will Brazil unban X before October?

\medskip
\noindent\textbf{Economics.}\\
-- Russian Ruble \textyen{}110 to \$1 USD by December 20?\\
-- U.S. Recession in 2024?

\medskip
\noindent\textbf{Elections.}\\
-- Will Stancil win Minnesota House Democratic Primary?\\
-- House control after 2024 election?

\medskip
\noindent\textbf{Entertainment.}\\
-- Will ``Oppenheimer'' win the Oscar for Best Picture?\\
-- Taylor Swift cancels more tour dates in August?

\medskip
\noindent\textbf{Financials.}\\
-- European Central Bank cuts rates in Oct meeting?\\
-- Trump vs.\ Biden: First to 1B?

\medskip
\noindent\textbf{Politics.}\\
-- Another Iran strike on Israel in 2024?\\
-- Saudi and Israel peace deal by March 2024?

\medskip
\noindent\textbf{Science/Technology.}\\
-- GPT-5 released in 2024?\\
-- Will Gabriel Haines win Crypto: The Game?
\end{ttfamily}
\end{shaded}

\medskip
\noindent\textbf{Realized counts per domain.} Recall set: Climate~100;
Commodities, Companies, Economics, Elections, Entertainment,
Financials, Politics, Science/Technology each~150. Total recall:
\textbf{1{,}300}. Forecast set: 150 per domain; total forecast:
\textbf{1{,}350}.

\subsection{Model Coverage and Knowledge Cutoffs}
\label{app:model-coverage}

We evaluate five frontier LLMs across two independent passes (recall
and forecast). Four of the five have graded recall scores and enter
the recall--forecast regression; the fifth (Gemini 3 Flash) appears
only in the forecast set and serves as a coverage cross-check. The
forecast window's earliest \texttt{close\_time} is 2026-09-01, which
sits 19--35 months past every model's stated knowledge cutoff,
substantially reducing direct contamination of the forecast pass.

\medskip
\noindent\textbf{Model identities and stated cutoffs.}

\begin{shaded}
\begin{ttfamily}
\small
\noindent\textbf{Claude Opus 4.7}\\
Provider: Anthropic.\quad
API ID: \texttt{claude-opus-4-7}.\\
Stated knowledge cutoff: early 2025.\\
Months from cutoff to earliest forecast close: $\approx$\,20.

\medskip
\noindent\textbf{Claude Sonnet 4.6}\\
Provider: Anthropic.\quad
API ID: \texttt{claude-sonnet-4-6}.\\
Stated knowledge cutoff: early 2025.\\
Months from cutoff to earliest forecast close: $\approx$\,20.

\medskip
\noindent\textbf{GPT-5.5}\\
Provider: OpenAI.\quad
API ID: \texttt{gpt-5.5}.\\
Stated knowledge cutoff: late 2024.\\
Months from cutoff to earliest forecast close: $\approx$\,22--24.

\medskip
\noindent\textbf{Gemini 3.1 Pro (preview)}\\
Provider: Google DeepMind.\quad
API ID: \texttt{gemini-3.1-pro-preview}.\\
Stated knowledge cutoff: early 2025.\\
Months from cutoff to earliest forecast close: $\approx$\,20.

\medskip
\noindent\textbf{Gemini 3 Flash (preview)}\\
Provider: Google DeepMind.\quad
API ID: \texttt{gemini-3-flash-preview}.\\
Stated knowledge cutoff: mid 2024.\\
Months from cutoff to earliest forecast close: $\approx$\,26.
\end{ttfamily}
\end{shaded}

\medskip
\noindent\textbf{Per-model graded-item counts.} Each model is queried
on every market in its assigned dataset; counts below are post-hoc
diagnostics rather than design choices. \emph{Recall items} = number of
recall-pass JSON entries the model produced. \emph{Recall graded} =
number with a non-null \texttt{recall\_score} after the two-stage
grading pipeline (~\ref{app:grader-prompt}). \emph{Brier-able}
= number of recall items with both a numeric \texttt{p\_yes} and a
binary \texttt{actual\_resolution} (used as the dependent variable
in the headline regression). \emph{Forecast items} = number of
forecast-pass JSON entries.

\begin{shaded}
\begin{ttfamily}
\small
\noindent\textbf{Claude Opus 4.7.}\\
recall items 1300\quad recall graded 891\quad Brier-able 1264\quad
forecast items 1313

\medskip
\noindent\textbf{Claude Sonnet 4.6.}\\
recall items 1300\quad recall graded 888\quad Brier-able 1264\quad
forecast items 1313

\medskip
\noindent\textbf{GPT-5.5.}\\
recall items 1300\quad recall graded 889\quad Brier-able 1105\quad
forecast items 1104

\medskip
\noindent\textbf{Gemini 3.1 Pro.}\\
recall items 1300\quad recall graded 957\quad Brier-able 1264\quad
forecast items 1308

\medskip
\noindent\textbf{Gemini 3 Flash.}\\
recall items 0\quad recall graded 0\quad Brier-able 0\quad
forecast items 1229
\end{ttfamily}
\end{shaded}

\textbf{Inclusion in the headline recall regression.} The
recall--forecast regression is fit over the four models with
non-empty graded-recall coverage: Claude Opus 4.7, Claude Sonnet 4.6,
GPT-5.5, and Gemini 3.1 Pro. This yields a panel of $4 \times 9 = 36$
(model, domain) cells at the aggregate level and $3 \times 1264 +
1 \times 1105 = 4{,}897$ (model, market) cells at the market level
(reduced to $4{,}897$ in the headline market-level spec after
restricting to rows with both a graded recall score and a Brier-able
target).

\subsection{Local Regression Specifications}
\label{app:regression-specifications}

\begin{figure}[ht]
  \begin{center}
    \centerline{\includegraphics[width=\columnwidth]{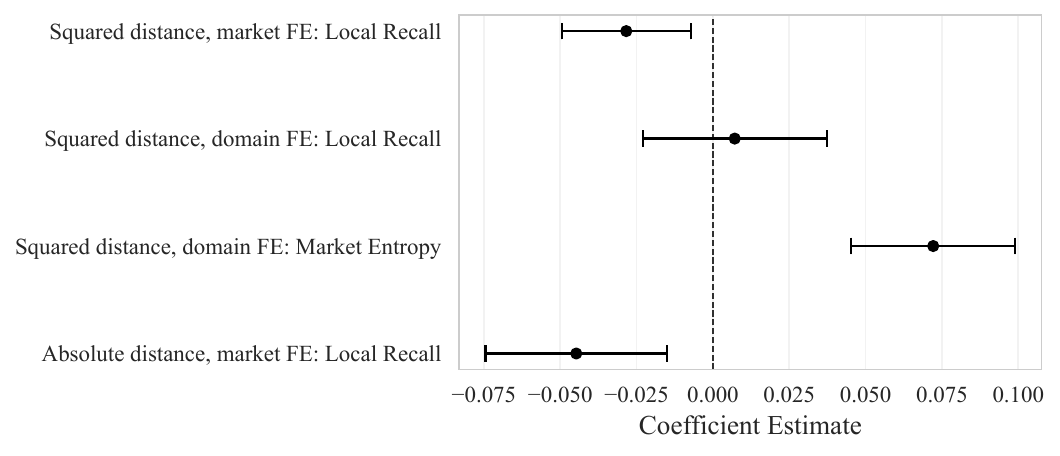}}
    \caption{Open-market proxy regression coefficients with 95\% confidence intervals by condition ID. Local historical recall is negatively associated with squared and absolute forecast-market disagreement in market fixed-effect specifications, while the domain-level specification shows that market entropy is strongly positively associated with forecast-market disagreement.}
    \label{app:open-proxy-coefficients}
  \end{center}
\end{figure}

Let $m$ index models, $i$ index resolved target markets, $j$ index
unresolved open markets, and $d(\cdot)$ denote domain membership. For
resolved markets, the dependent variable is the Brier loss
\begin{equation}
B_{m,i}=(p_{m,i}-y_i)^2,
\end{equation}
where $p_{m,i}$ is model $m$'s probability assigned to the ``Yes''
outcome and $y_i \in \{0,1\}$ is the realized resolution. The primary
market-level specification is
\begin{equation}
B_{m,i}
=
\alpha_m
+
\lambda_i
+
\beta R^{\mathrm{local}}_{m,i,10}
+
\varepsilon_{m,i},
\label{eq:app-market-fe}
\end{equation}
where $\alpha_m$ are model fixed effects, $\lambda_i$ are market fixed
effects, and $R^{\mathrm{local}}_{m,i,10}$ is the mean recall-fidelity
score over the ten nearest resolved historical neighbors of market $i$.
The weighted-neighbor robustness check uses the same specification but
replaces $R^{\mathrm{local}}_{m,i,10}$ with the cosine-similarity
weighted score $R^{\mathrm{local},w}_{m,i,10}$. Standard errors are
clustered by condition ID in all market-level regressions.

For aggregate domain-level analyses, let
\[
\bar{B}_{m,d}
=
\frac{1}{|\mathcal{I}_d|}
\sum_{i \in \mathcal{I}_d}
B_{m,i},
\qquad
R_{m,d}
=
\frac{1}{|\mathcal{I}_d|}
\sum_{i \in \mathcal{I}_d}
R_{m,i},
\]
where $\mathcal{I}_d$ is the set of resolved markets in domain $d$.
No-recall/null recall observations are treated as $0.0$ when computing
$R_{m,d}$. Domain difficulty is the mean binary entropy of Polymarket
prices over open markets in the same domain,
\begin{equation}
D_d
=
\frac{1}{|\mathcal{O}_d|}
\sum_{j \in \mathcal{O}_d}
\left[
-\pi_j \log(\pi_j)
-
(1-\pi_j)\log(1-\pi_j)
\right],
\end{equation}
where $\pi_j$ is the contemporaneous market-implied probability of
``Yes'' for open market $j$. The full domain-level fixed-effect
specification is
\begin{equation}
\bar{B}_{m,d}
=
\alpha_m
+
\beta R_{m,d}
+
\gamma D_d
+
\varepsilon_{m,d}.
\label{eq:app-domain-fe}
\end{equation}
The bivariate and difficulty-controlled variants omit $\alpha_m$ and/or
$D_d$ as indicated in Table~\ref{tab:app-historical-regressions}.

\begin{table}[t]
\centering
\small
\setlength{\tabcolsep}{4pt}
\resizebox{\textwidth}{!}{%
\begin{tabular}{lllcccc}
\toprule
Panel & Outcome & Main regressor & Controls / FE & $\hat{\beta}$ (SE) & $p$ & $N$ / clusters \\
\midrule
Historical local & $B_{m,i}$ & $R^{\mathrm{local}}_{m,i,10}$ &
Model + market FE & $-0.0652$ $(0.0231)$ & $0.0048$ & $4{,}897 / 1{,}264$ \\
Historical weighted & $B_{m,i}$ & $R^{\mathrm{local},w}_{m,i,10}$ &
Model + market FE & $-0.0710$ $(0.0225)$ & $0.0016$ & $4{,}897 / 1{,}264$ \\
Domain M1 & $\bar{B}_{m,d}$ & $R_{m,d}$ &
None & $-0.2038$ $(0.0430)$ & $<0.001$ & $36$ / -- \\
Domain M2 & $\bar{B}_{m,d}$ & $R_{m,d}$ &
$D_d$ & $-0.2216$ $(0.0445)$ & $<0.001$ & $36$ / -- \\
Domain M3 & $\bar{B}_{m,d}$ & $R_{m,d}$ &
Model FE + $D_d$ & $-0.2227$ $(0.0361)$ & $<0.001$ & $36$ / -- \\
\bottomrule
\end{tabular}
}
\caption{Historical resolved-market and aggregate domain-level recall regressions. The reported coefficient is always the coefficient on the recall-fidelity regressor. In Domain M2, $\hat{\gamma}_{D}=-0.1185$ ($SE=0.0891$, $p=0.193$); in Domain M3, $\hat{\gamma}_{D}=-0.1192$ ($SE=0.0691$, $p=0.0948$).}
\label{tab:app-historical-regressions}
\end{table}

For unresolved open markets, true resolutions are unavailable. We
therefore use market-price proxy outcomes:
\[
\widetilde{B}_{m,j}=(p_{m,j}-\pi_j)^2,
\qquad
\widetilde{A}_{m,j}=|p_{m,j}-\pi_j|,
\]
and the soft-target log loss
\[
\widetilde{L}_{m,j}
=
-\pi_j\log(p_{m,j})
-
(1-\pi_j)\log(1-p_{m,j}).
\]
The market fixed-effect open-market specification is
\begin{equation}
Y_{m,j}
=
\alpha_m
+
\lambda_j
+
\beta R^{\mathrm{local}}_{m,j,10}
+
\varepsilon_{m,j},
\label{eq:app-open-market-fe}
\end{equation}
where $Y_{m,j}\in\{\widetilde{B}_{m,j},\widetilde{A}_{m,j},
\widetilde{L}_{m,j}\}$. The domain-level open-market comparison is
\begin{equation}
Y_{m,j}
=
\alpha_m
+
\delta_{d(j)}
+
\beta R^{\mathrm{local}}_{m,j,10}
+
\gamma H(\pi_j)
+
\varepsilon_{m,j},
\label{eq:app-open-domain-fe}
\end{equation}
where $H(\pi_j)$ is the binary entropy of the market price. Standard
errors are again clustered by condition ID.

\begin{table}[t]
\centering
\small
\setlength{\tabcolsep}{4pt}
\resizebox{\textwidth}{!}{%
\begin{tabular}{lllcccc}
\toprule
Panel & Outcome & Term & Controls / FE & $\hat{\beta}$ (SE) & $p$ & $N$ / clusters \\
\midrule
Open market FE & $\widetilde{B}_{m,j}$ & $R^{\mathrm{local}}_{m,j,10}$ &
Model + market FE & $-0.0284$ $(0.0108)$ & $0.0084$ & $5{,}724 / 1{,}897$ \\
Open market FE & $\widetilde{A}_{m,j}$ & $R^{\mathrm{local}}_{m,j,10}$ &
Model + market FE & $-0.0448$ $(0.0152)$ & $0.0032$ & $5{,}724 / 1{,}897$ \\
Open market FE & $\widetilde{L}_{m,j}$ & $R^{\mathrm{local}}_{m,j,10}$ &
Model + market FE & $-0.0606$ $(0.0696)$ & $0.384$ & $5{,}724 / 1{,}897$ \\
Open domain FE & $\widetilde{B}_{m,j}$ & $R^{\mathrm{local}}_{m,j,10}$ &
Model + domain FE + $H(\pi_j)$ & $0.0071$ $(0.0154)$ & $0.642$ & $5{,}724 / 1{,}897$ \\
Open domain FE & $\widetilde{B}_{m,j}$ & $H(\pi_j)$ &
Model + domain FE & $0.0721$ $(0.0137)$ & $<0.001$ & $5{,}724 / 1{,}897$ \\
Open domain FE & $\widetilde{L}_{m,j}$ & $R^{\mathrm{local}}_{m,j,10}$ &
Model + domain FE + $H(\pi_j)$ & $0.0140$ $(0.0544)$ & $0.797$ & $5{,}724 / 1{,}897$ \\
Open domain FE & $\widetilde{L}_{m,j}$ & $H(\pi_j)$ &
Model + domain FE & $1.2601$ $(0.0469)$ & $<0.001$ & $5{,}724 / 1{,}897$ \\
\bottomrule
\end{tabular}
}
\caption{Open-market proxy regressions. Market fixed effects compare models on the same unresolved market. Domain fixed effects with entropy controls test whether local recall explains broader domain-level variation after accounting for market-price uncertainty.}
\label{tab:app-open-regressions}
\end{table}

\subsection{Market Entropy}
\label{app:entropy_defn}
For a binary market with market probability $p_m$, we define:
\begin{equation}
\label{eq:entropy}
H(p_m)
=
-
p_m \log(p_m)
-
(1-p_m)\log(1-p_m).
\end{equation}
{\small If $p_m \approx 0.5$, entropy is high because the market is uncertain. If $p_m \approx 0$ or $p_m \approx 1$, entropy is low because the market is confident.}

\subsection{Modelwise, Domainwise Metrics}
\label{app:modelwise-domainwise}
\begin{center}
\begin{small}
\setlength{\tabcolsep}{4pt}
\begin{tabular}{llccccc}
\toprule
Domain & Metric & Opus 4.7 & Sonnet 4.6 & Gemini 3.1 Pro & GPT-5.5 & Overall \\
\midrule

\multirow{4}{*}{Climate}
& FC Loss      & 0.0534 & 0.0598 & 0.0540 & 0.0486 & 0.0540 \\
& FC Log Loss  & 0.5714 & 0.5971 & 0.5645 & 0.5714 & 0.5761 \\
& RC Brier     & 0.1271 & 0.1452 & 0.1139 & 0.1425 & 0.1322 \\
& RC Fidelity  & 0.9167 & 0.8864 & 1.0000 & 0.8145 & 0.9044 \\
\midrule

\multirow{4}{*}{Commodities}
& FC Loss      & 0.0911 & 0.0858 & 0.0802 & 0.0666 & 0.0809 \\
& FC Log Loss  & 0.7709 & 0.7604 & 0.7225 & 0.6932 & 0.7367 \\
& RC Brier     & 0.0978 & 0.1236 & 0.0843 & 0.0747 & 0.0951 \\
& RC Fidelity  & 0.8488 & 0.8645 & 0.8704 & 0.8380 & 0.8554 \\
\midrule

\multirow{4}{*}{Companies}
& FC Loss      & 0.1139 & 0.1148 & 0.1156 & 0.1135 & 0.1144 \\
& FC Log Loss  & 0.7413 & 0.7547 & 0.7366 & 0.7560 & 0.7472 \\
& RC Brier     & 0.1092 & 0.1507 & 0.1120 & 0.0941 & 0.1165 \\
& RC Fidelity  & 0.8814 & 0.8626 & 0.9143 & 0.8820 & 0.8851 \\
\midrule

\multirow{4}{*}{Economics}
& FC Loss      & 0.0708 & 0.0639 & 0.0643 & 0.0531 & 0.0631 \\
& FC Log Loss  & 0.6553 & 0.6353 & 0.6358 & 0.5734 & 0.6249 \\
& RC Brier     & 0.0666 & 0.0855 & 0.0578 & 0.0516 & 0.0654 \\
& RC Fidelity  & 0.9651 & 0.9609 & 0.9733 & 0.9457 & 0.9612 \\
\midrule

\multirow{4}{*}{Elections}
& FC Loss      & 0.0457 & 0.0581 & 0.0612 & 0.0393 & 0.0511 \\
& FC Log Loss  & 0.5119 & 0.5343 & 0.5317 & 0.4595 & 0.5094 \\
& RC Brier     & 0.0777 & 0.1063 & 0.0551 & 0.0567 & 0.0740 \\
& RC Fidelity  & 0.9573 & 0.9043 & 0.9606 & 0.8740 & 0.9241 \\
\midrule

\multirow{4}{*}{Entertainment}
& FC Loss      & 0.0600 & 0.0906 & 0.0680 & 0.0651 & 0.0709 \\
& FC Log Loss  & 0.7020 & 0.8191 & 0.7138 & 0.7244 & 0.7399 \\
& RC Brier     & 0.0867 & 0.1338 & 0.0584 & 0.0535 & 0.0831 \\
& RC Fidelity  & 0.9423 & 0.9100 & 0.9250 & 0.9181 & 0.9239 \\
\midrule

\multirow{4}{*}{Financials}
& FC Loss      & 0.0776 & 0.0760 & 0.0957 & 0.0780 & 0.0818 \\
& FC Log Loss  & 0.6817 & 0.6825 & 0.7221 & 0.7566 & 0.7107 \\
& RC Brier     & 0.1355 & 0.1925 & 0.0947 & 0.0984 & 0.1303 \\
& RC Fidelity  & 0.8483 & 0.8224 & 0.9069 & 0.8663 & 0.8610 \\
\midrule

\multirow{4}{*}{Science/Technology}
& FC Loss      & 0.0642 & 0.0871 & 0.0934 & 0.0621 & 0.0767 \\
& FC Log Loss  & 0.6075 & 0.6912 & 0.6695 & 0.5667 & 0.6337 \\
& RC Brier     & 0.1223 & 0.1737 & 0.1321 & 0.1096 & 0.1344 \\
& RC Fidelity  & 0.9433 & 0.8978 & 0.9588 & 0.9111 & 0.9278 \\
\midrule

\multirow{4}{*}{Overall}
& FC Loss      & 0.0721 & 0.0795 & 0.0791 & 0.0658 & 0.0741 \\
& FC Log Loss  & 0.6552 & 0.6843 & 0.6621 & 0.6377 & 0.6598 \\
& RC Brier     & 0.1029 & 0.1389 & 0.0885 & 0.0851 & 0.1039 \\
& RC Fidelity  & 0.9129 & 0.8886 & 0.9386 & 0.8812 & 0.9053 \\

\bottomrule
\end{tabular}
\end{small}
\end{center}

\subsection{Murphy Breakdown}
\label{app:murphy}
\begin{figure}[ht]
  \begin{center}
    \centerline{\includegraphics[width=\columnwidth]{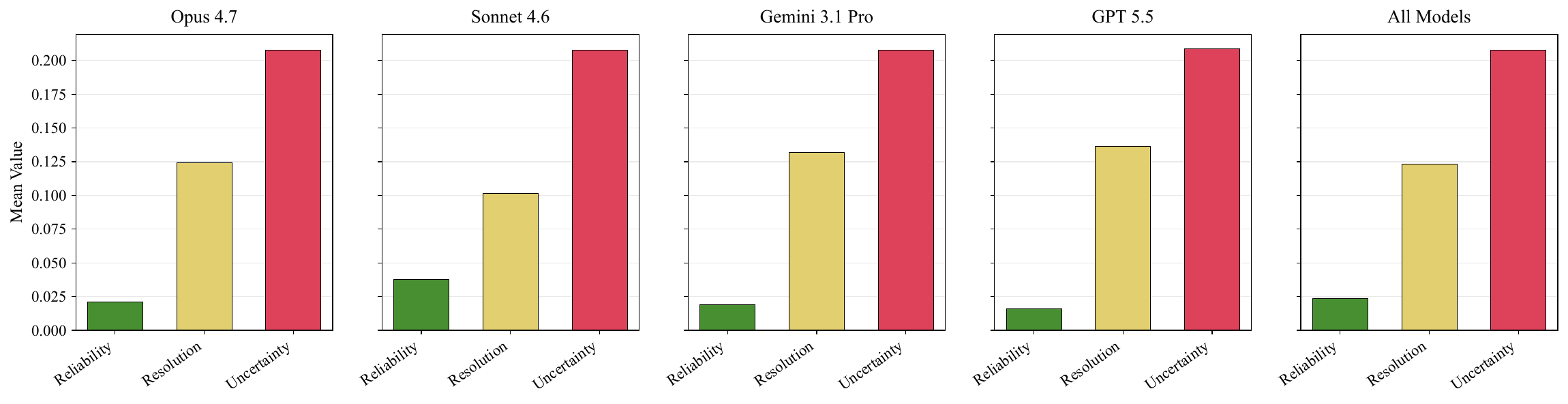}}
    \caption{Murphy Breakdown of recall Brier for each model. A lower reliability score is more favorable while a higher resolution score is more favorable.}
    \label{fig: murphy-breakdown-by-model}
  \end{center}
\end{figure}

\begin{table}[t]
\centering
\small
\begin{tabular}{lcccccc}
\toprule
Model & $n$ & Brier & Uncertainty & Reliability & Resolution & Null Rate \\
\midrule
Claude Opus 4.7     & 1150 & 0.1017 & 0.2110 & 0.0102 & 0.0031 & 0.3174 \\
Claude Sonnet 4.6   & 1150 & 0.1406 & 0.2110 & 0.0239 & 0.0052 & 0.3252 \\
Gemini 3.1 Pro      & 1150 & 0.0877 & 0.2110 & 0.0080 & 0.0042 & 0.2591 \\
GPT-5.5             & \phantom{0}993 & 0.0816 & 0.2132 & 0.0075 & 0.0022 & 0.2115 \\
\bottomrule
\end{tabular}
\caption{Murphy decomposition components and null prediction rates by model. Lower values are better for Brier, Reliability, and Null Rate, while higher values are better for Resolution.}
\label{tab:murphy_decomp}
\end{table}

\end{document}